\pdfoutput=1

\documentclass[11pt]{article}

\usepackage[preprint]{acl}

\usepackage{times}
\usepackage{latexsym}
\usepackage[T1]{fontenc}
\usepackage[utf8]{inputenc}
\usepackage{microtype}
\usepackage{inconsolata}
\usepackage{graphicx}
\usepackage{amsmath} 
\usepackage{algorithm}
\usepackage{algorithmic}
\usepackage{float}
\usepackage{booktabs}    
\usepackage{multirow}    
\usepackage{array}       
\usepackage{amsfonts}  
\usepackage{amssymb}   
\usepackage{dsfont}
\usepackage{tabularx}   
\usepackage{makecell}   
\usepackage{algorithm}
\usepackage{algorithmic}
\usepackage{mathtools}
\usepackage{enumitem}
\usepackage{footmisc}
\usepackage[table]{xcolor}
\definecolor{seclight}{gray}{0.92}       
\definecolor{sxblue}{RGB}{229,241,255}   

\newcommand{\Autoref}[1]{%
  \begingroup%
  \def\chapterautorefname{Chapter}%
  \def\sectionautorefname{Section}%
  \def\subsectionautorefname{Section}%
  \def\subsubsectionautorefname{Section}%
  \def\paragraphautorefname{Paragraph}%
  \def\tableautorefname{Table}%
  \def\equationautorefname{Equation}%
  \def\algorithmautorefname{Algorithm}%
  \autoref{#1}%
  \endgroup%
}

\usepackage[most]{tcolorbox}
\tcbuselibrary{skins, breakable}
\tcbset{
  aclbox/.style={
    colback=gray!5,
    colframe=black!30,
    boxrule=0.5pt,
    arc=2pt,
    left=2pt,right=2pt,top=2pt,bottom=2pt,
  }
}

\title{Semantic XPath: Structured Agentic Memory Access for Conversational AI}

\author{
  Yifan Simon Liu\thanks{Equal contribution}\textsuperscript{\textnormal{1}},
  Ruifan Wu\footnotemark[1]\textsuperscript{\textnormal{1}},
  Liam Gallagher\footnotemark[1]\textsuperscript{\textnormal{1}},
  Jiazhou Liang\footnotemark[1]\textsuperscript{\textnormal{1}}\\
  \textbf{Armin Toroghi}\textsuperscript{\textnormal{1}},
\textbf{Scott Sanner}\textsuperscript{\textnormal{1,2}}\\
  \textsuperscript{1}University of Toronto, Canada\\
  \textsuperscript{2}Vector Institute of Artificial Intelligence, Toronto, Canada
}

\begin{document}
\makeatletter
\acl@anonymizefalse
\makeatother
\maketitle

\begin{abstract}
Conversational AI (ConvAI) agents increasingly maintain structured memory to support long-term, task-oriented interactions. In-context memory approaches append the growing history to the model input, which scales poorly under context-window limits. RAG-based methods retrieve request-relevant information, but most assume flat memory collections and ignore structure.
We propose \textsc{Semantic XPath}, a tree-structured 
memory module to access and update structured conversational memory. 
\textsc{Semantic XPath} improves performance over flat-RAG baselines by 176.7\% while using only 9.1\% of the tokens required by in-context memory. We also introduce \textsc{SemanticXPath Chat}, an end-to-end ConvAI demo system that visualizes the structured memory and query execution details. Overall, this paper demonstrates a candidate for the next generation of long-term, task-oriented ConvAI systems built on structured memory. 
\end{abstract}

\section{Introduction}

Conversational AI (ConvAI) agents aim to support long-term, task-oriented interactions in which users manage multiple tasks (e.g., travel itinerary planning, to do list management, meal kit recipe recommendation) and repeatedly inspect and revise their evolving artifacts across many versions. To enable such interactions, ConvAI agents increasingly maintain \emph{structured memory} that preserves the conversation history, revision history, and hierarchical organization of each task artifact (\citealt{memtree, hmem}).

\begin{figure}
    \centering
    \includegraphics[width=\linewidth]{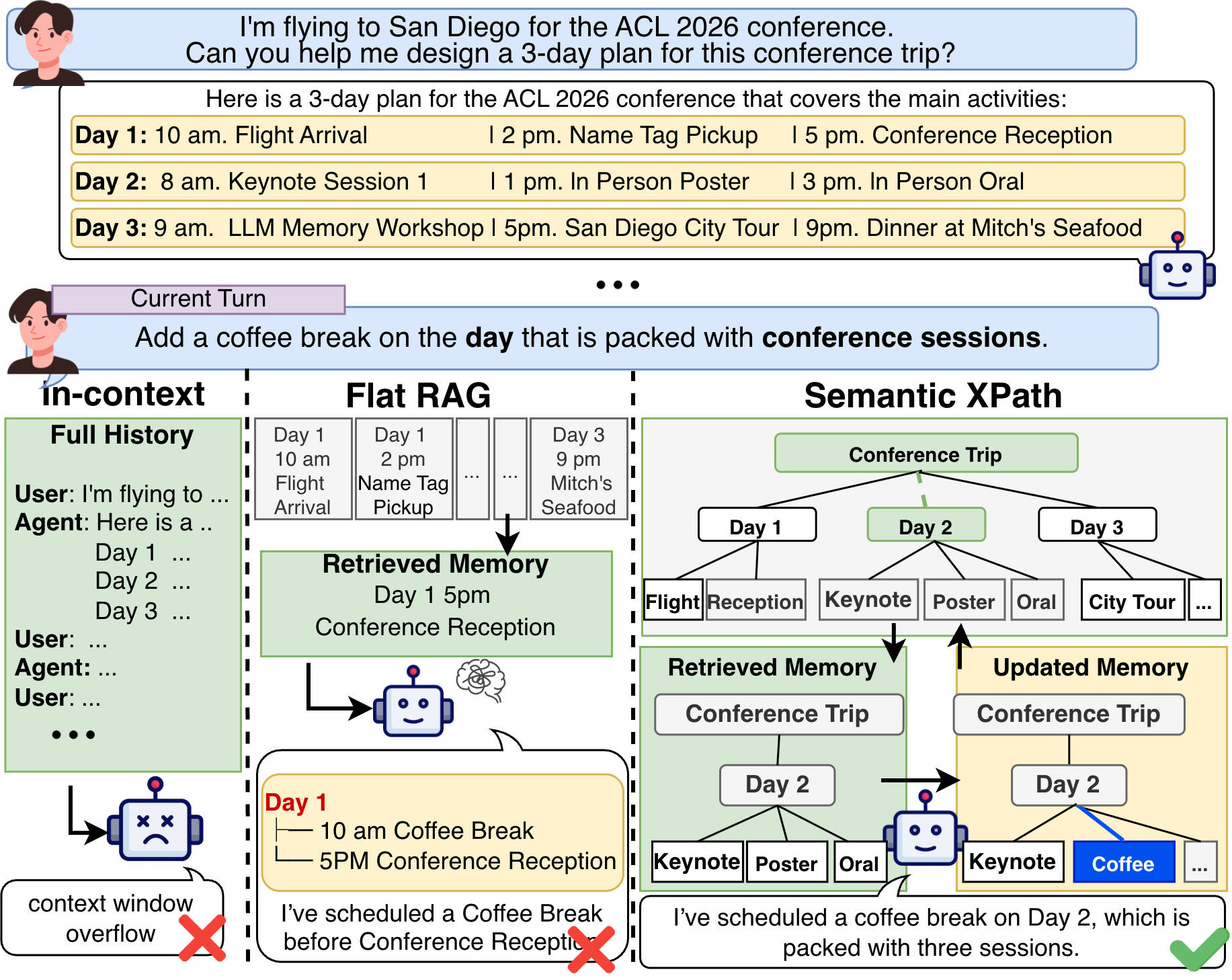}
    \caption{A travel-planning ConvAI agent for a 3-day ACL conference trip in San Diego. \textbf{Left:} in-context memory appends the full conversation. \textbf{Middle:} flat RAG retrieves from a flattened collection of memory items. \textbf{Right:} \textsc{Semantic XPath} retrieves and updates the relevant substructure from structured memory.}
    \label{fig:example}
\end{figure}

\begin{figure*}
    \centering
    \includegraphics[width=1.0\linewidth]{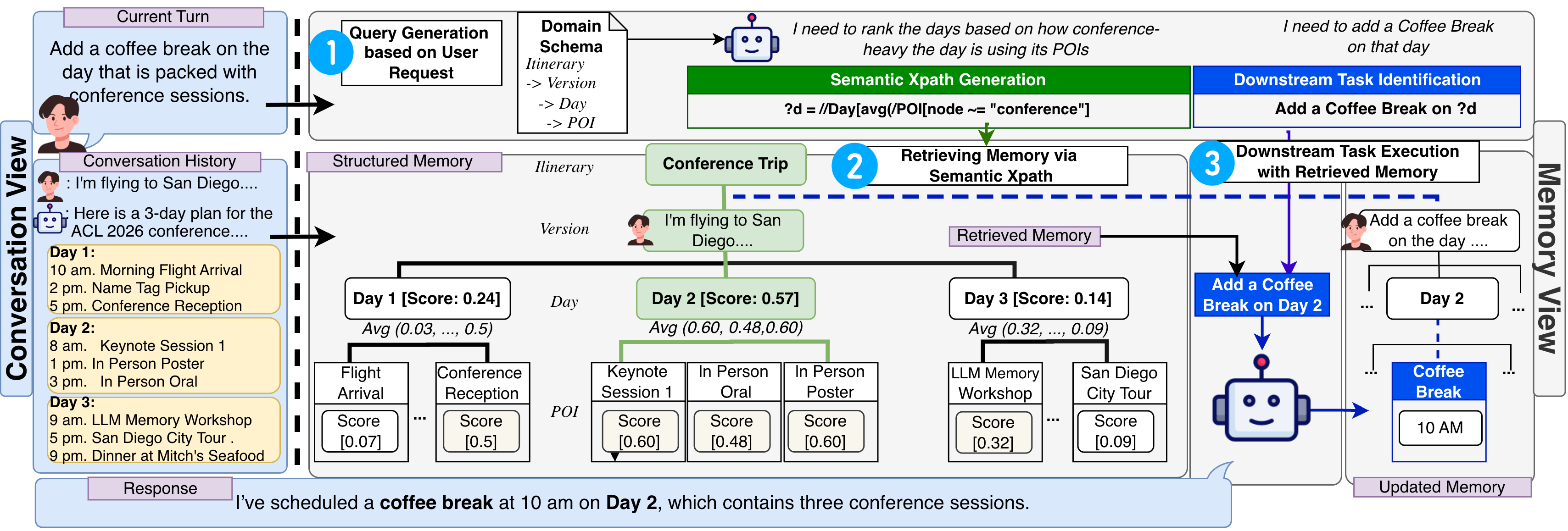}
    \caption{End-to-end \textsc{Semantic XPath} pipeline on an ACL 2026 trip plan illustrated in both the conversation view (blue) and the memory view (grey). \textbf{Step 1:} The user requests to \emph{``add a coffee break on the day that is packed with conference sessions.''} is translated into a \textsc{Semantic XPath} query based on the memory schema. \textbf{Step 2:} The query is executed with structural matching and semantic relevance scoring, selecting \emph{Day~2}. \textbf{Step 3:} The retrieved substructure is passed to downstream generation to insert a coffee-break POI and respond to the user.}
    \label{fig:pipeline}
\end{figure*}

Early ConvAI agents rely on an in-context approach that appends the growing conversation history to the model input (\Autoref{fig:example}, left; \citealt{dialoGPT}). However, this approach scales poorly under context-window limits and incurs higher token costs and latency. Long contexts also lead to poor reasoning and hallucinated outputs, thus reducing reliability \citep{lost_middle, selective_attn}. 
Our empirical results show that the in-context memory approach fails to satisfy half of user requests after only five interaction turns.

To address these limitations, RAG-based 
systems have been proposed to retrieve information relevant to the current user request \citep{rag, memgpt}. Many of these approaches treat memory as a flat collection of items during retrieval, which ignores natural hierarchical structure often present in memory \citep{memtree, raptor}. For example, in a 3-day ACL trip plan, when a user asks to \emph{“add a coffee break on the day that is packed with conference sessions,”} flat RAG approaches may retrieve conference-related activities from the wrong day due to a lack of structured \mbox{memory organization (\Autoref{fig:example}, middle).}

These limitations motivate a structure-aware RAG method that retrieves relevant structured data from memory. A well-known solution for structure-aware retrieval in hierarchical tree-structured data is the XPath query language \citep{w3c-xpath10}, but it lacks semantic awareness and relies on exact matches. Therefore, in this paper:
\begin{enumerate}
[leftmargin=*, itemsep=2pt, topsep=2pt]
    \item We propose \textsc{Semantic XPath} (\Autoref{fig:example}, right), a tree-structured memory module that uses an XPath-style query language to retrieve only the relevant memory substructure for efficient memory access and updates.
        
    \item We empirically show that \textsc{Semantic XPath} improves performance over flat-RAG baselines by 176.7\% while using only 9.1\% of the tokens required by in-context memory.
    \item We introduce \textsc{SemanticXPath Chat}\footnote{\label{fn:demo} Try live demo at \url{https://semanticxpathchat.com}.}, an end-to-end ConvAI system for long-term, task-oriented interaction that demonstrates \textsc{Semantic XPath} with visualizations of the structured memory and query execution details.
\end{enumerate}
\section{System Description}\label{sec:sys_desc}
\begin{figure*}[t]
    \centering
    \includegraphics[width=0.95\textwidth]{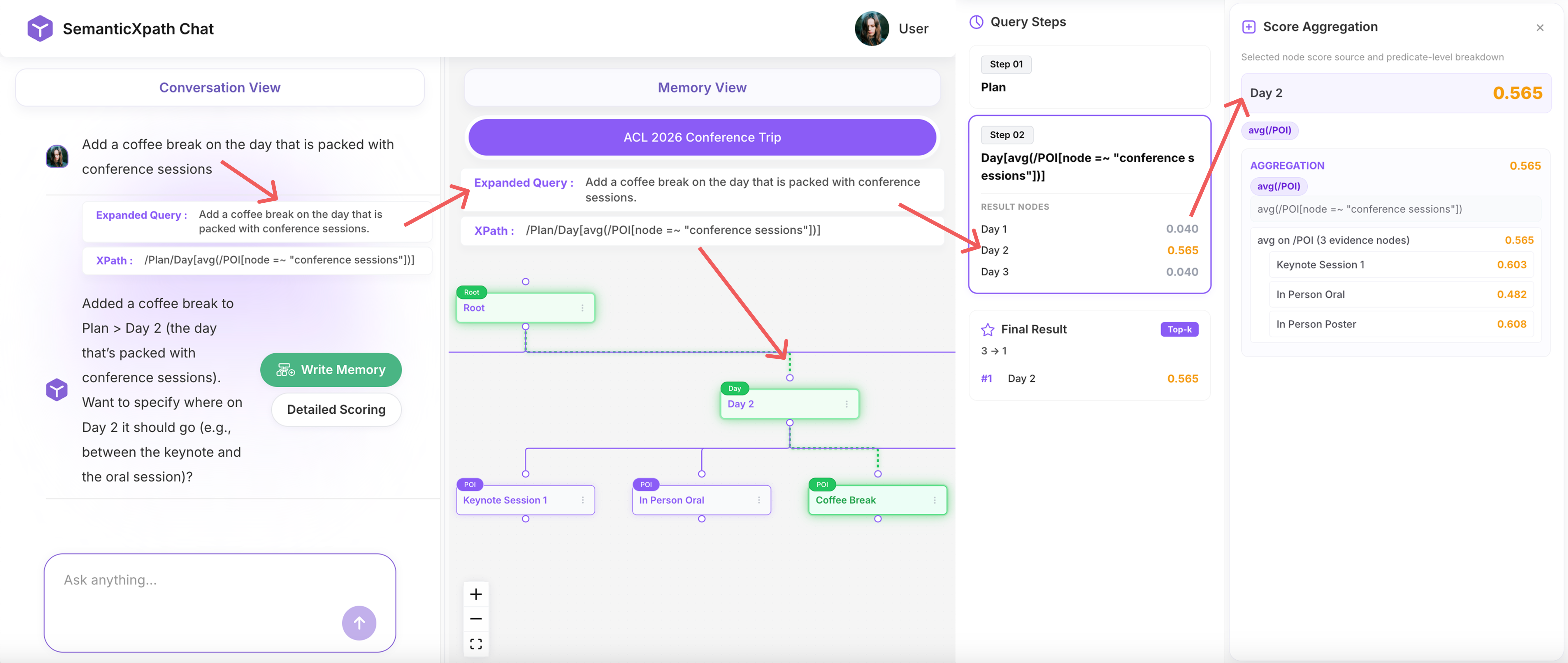}
    \caption{\textsc{SemanticXPath Chat} demonstration\protect\footref{fn:demo}. \textbf{Left:} Conversation view, where the user asks to \textit{add a coffee break on the day packed with conference sessions}. \textbf{Middle:} Memory view, highlighting the retrieved \emph{Day~2} and the newly inserted coffee break. \textbf{Right:} Execution view, showing step-by-step query execution and scoring details.}
    \label{fig:demo}
\end{figure*}

\paragraph{Overview.}
To support long-term, task-oriented interaction between users and agents, we introduce \textsc{SemanticXPath Chat}, an end-to-end ConvAI system that maintains task-specific, tree-structured memory. This system uses a tree-structured memory module supported by \textsc{Semantic XPath} to retrieve only the relevant structured data at the appropriate level of granularity, which in turn enables efficient memory access and updates.


\Autoref{fig:pipeline} shows an example workflow. After creating a 3-day ACL conference trip, the user asks to “\emph{add a coffee break on the day packed with conference sessions}.” The system first translates the request into a \textsc{Semantic XPath} query grounded in the memory schema, then executes it to rank candidates. The top-ranked result is \emph{Day~2}, which contains three conference-related activities. The system then takes the retrieved node and the corresponding subtree for downstream generation, inserts a coffee-break activity into \emph{Day~2}, and records the update by creating a new version branch in structured memory.

\paragraph{Features.}
\Autoref{fig:demo} shows a screenshot from the \textsc{SemanticXPath Chat} system to illustrate key features on the same running example.  We refer the reader to the demonstration video\footnote{\url{https://www.youtube.com/watch?v=l-DrEr4P5JA}} for a detailed walkthrough of the system.

 \textbf{Conversation View (left)} shows the conversation history. Each turn includes action buttons bringing the user to the structured memory view and the execution view in the right panels.

\textbf{Memory View (middle)} visualizes the structured memory. The highlighted path indicates the relevant path for the current user request. (e.g., \emph{Day~2} and added  coffee break are highlighted.)

\textbf{Execution View (right)} visualizes the step-by-step execution and socring. Users can select a query step to inspect its execution details and click candidate nodes to view semantic relevance scores. We present \textsc{Semantic XPath} query language and execution details in \Autoref{sec:method}.

\section{Semantic XPath}\label{sec:xpath}
\label{sec:method}

\textsc{Semantic XPath} is a tree-structured memory module that uses an XPath-style query language to retrieve only the relevant structured data for efficient memory access and updates. \Autoref{fig:pipeline} illustrates the overall pipeline.

\subsection{Data Model}

We represent structured conversation memory as a rooted tree
\[
T = (V, E, r),
\]
where each node $v \in V$ corresponds to a unit of conversational
state, $E$ denotes parent--child relations, and $r \in V$ is the root. Each node is associated with a node type and a set of textual attributes associated with the node.

The data model is governed by a schema derived from the natural hierarchy of the conversation and task information, which specifies the node types and their attributes.


\begin{tcolorbox}[aclbox, width=\linewidth]
\textbf{ACL Trip Example:} The schema for \textit{``3-day ACL conference trip''} example
(\Autoref{fig:pipeline}) is:
\begin{center}
\emph{Itinerary} $\rightarrow$ \emph{Version} $\rightarrow$ \emph{Day} $\rightarrow$ \emph{POI}.
\end{center}

\begin{itemize}[leftmargin=*, itemsep=1pt, topsep=1pt]
\item \emph{Itinerary} is the task-instance root.
\item \emph{Version} captures revision history (e.g., the initial plan and an updated version after adding a coffee break).
\item \emph{Day} and \emph{POI} are task-specific. Each day contains conference activities (e.g., \textit{``welcome reception''}).
\end{itemize}
\end{tcolorbox}


\subsection{Query Grammar}

\begin{tcolorbox}[title={Semantic XPath Grammar}, colback=white, colframe=black,     left=2pt,right=2pt,top=2pt,bottom=0pt]

\textbf{Query} \\
\quad $Q ::= \varnothing \mid S\, \varnothing \mid S \; Q$

\medskip
\textbf{Step} \\
\quad $S ::= A\,N \, [R] \, [P]$

\medskip
\textbf{Axis navigator} \\
\quad $A ::= \texttt{/} \mid \texttt{//}$

\medskip
\textbf{Node selector} \\
\quad $N ::= \texttt{NodeType} \mid *$

\medskip
\textbf{Positional selector} \\
\quad $R ::= [i] \mid [-i] \mid [i{:}j]$

\medskip
\textbf{Semantic Relevance Operator} \\
\quad $P ::= \frac{(P+P)}{2} \mid 
        P \cdot P 
      \mid \min(P, P)
      \mid \max(P,P)
      \mid 1-P 
      \mid \texttt{Local} 
      \mid \texttt{Agg}(S)
$

\medskip
\textbf{Local Semantic Relevance} \\
\quad $\texttt{Local} ::= 
      [\texttt{attr\_name} \approx \texttt{String}]
      \mid [\texttt{node} \approx \texttt{String}]$

\medskip
\textbf{Aggregation Semantic Relevance} \\
\quad $\texttt{Agg} ::= \texttt{avg} \mid \texttt{min} \mid \texttt{max} \mid \texttt{gmean}$

\label{box:semantic-xpath-grammar}
\end{tcolorbox}
A query \(Q\) (symbol \colorbox{gray!20}{\texttt{$Q$}}) is a sequence of steps. Each step (symbol \colorbox{gray!20}{\texttt{$S$}}) 
contains four components:

\begin{itemize}
    \item An axis navigator (symbol \colorbox{gray!20}{\texttt{$A$}}) that determines how the structured memory is traversed. \texttt{/} selects children and \texttt{//} selects descendants.
    
    \item A node selector
    (symbol \colorbox{gray!20}{\texttt{$N$}}), 
    which specifies either a concrete node type 
    or the wildcard.

    \item An optional positional selector
    (symbol \colorbox{gray!20}{\texttt{$R$}}).

    \item An optional semantic relevance operator
    (symbol \colorbox{gray!20}{\texttt{$P$}}), which assigns a semantic relevance score to matched nodes.
\end{itemize}

We walk through two concrete examples.
\begin{tcolorbox}[aclbox, width=\linewidth]
\textbf{ACL Trip Example (cont.):} 
The query $Q_1$ for a request \emph{``add a coffee break on the day packed with conference sessions''}
can be translated into:
\begin{center}
\path{//Day[ avg(/POI[ node ~= "conference"])]}
\end{center}
\end{tcolorbox}

$Q_1$ first matches all \emph{Day} nodes at any depth ($A=\textit{//}$, $N=\textit{Day}$). To reflect the notion of a day being \textit{``packed''} with conference sessions, it scores each day with an \emph{ aggregation semantic relevance operator} $\textit{Agg}(S')$, which aggregates POI-level local evidence $S'=\textit{POI}[\textit{Local}]$. Using \textit{avg} yields a density-style measure of how conference-heavy the day is.

\begin{tcolorbox}[aclbox, width=\linewidth]
\textbf{ACL Trip Example (cont.).}
The query $Q_2$ for \emph{``On the third day, remove all activities except the workshop.''}
can be translated into:
\begin{center}
\path{//Day[3]/ POI[1 - [node ~= "workshop"]]}
\end{center}
\end{tcolorbox}

$Q_2$ first uses a \emph{ positional selector}, \path{//Day[3]}, to match the third day in the structured memory. It then applies a \emph{local semantic relevance operator} at the \emph{POI} level (\textit{Local} = [node $\approx$ ``workshop'']), to score each POI by how closely its description matches \textit{``workshop''}. Finally, the \emph{semantic relevance operator} $1 - P$ then inverts this score to target all POIs that do \emph{not} match ``workshop'', identifying the activities to be removed.

\subsection{Query Execution}
\label{sec:query_exec}

\textsc{Semantic XPath} query execution maintains a weighted node set
$
W \;\subseteq\; V \times [0,1]
$,
where each pair $(u,w)$ assigns a weight $w$ to a node $u\in V$. 

\paragraph{Evaluation function.}
A query is executed by a single recursive evaluation function
$
\mathsf{E}:\mathcal{X}\times\mathcal{W}\rightarrow\mathcal{W},
$
where $\mathcal{X}$ ranges over well-formed syntax objects induced by the grammar
(e.g., queries, steps, and step components).

Execution is initialized at the root as $\mathsf{E}(Q,\{(r,1)\})$.
Query-level recursion is then:
\[
\mathsf{E}(Q,W)
=
\begin{cases}
W, & \text{if } Q=\varnothing,\\[6pt]
\mathsf{E}(S,W), & \text{if } Q=S\varnothing,\\[6pt]
\mathsf{E}\big(Q',\,\mathsf{E}(S,W)\big), & \text{if } Q=SQ'.
\end{cases}
\]
The remaining cases for the step $S$ and step components ($A$, $N$, $R$, $P$) are defined in \Autoref{app:eval}.

\begin{figure*}[t]
    \centering
    \includegraphics[width=\linewidth]{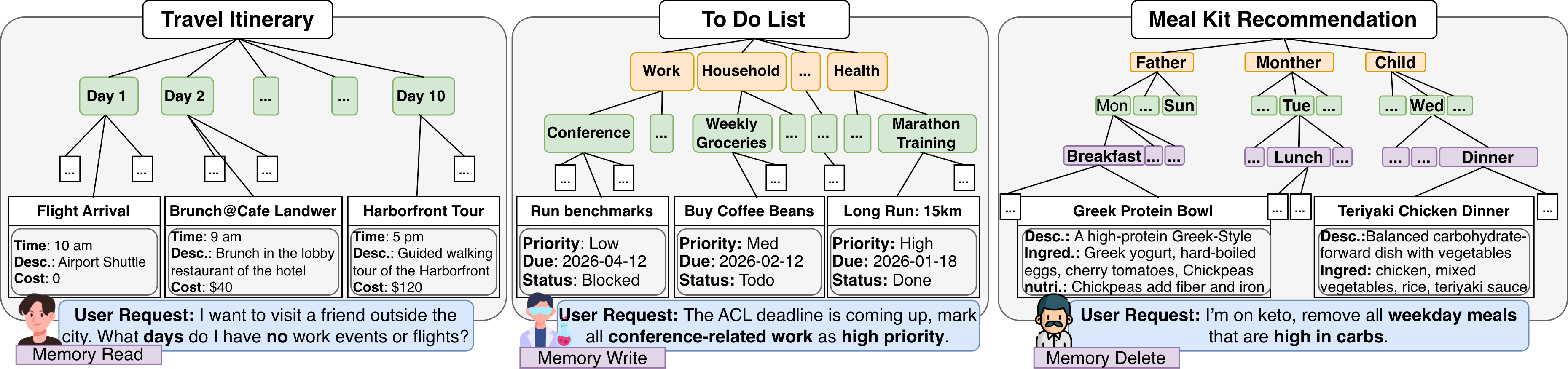}
    \caption{Example of structured memory for three ConvAI scenarios: Travel Itinerary, To Do List, and Meal Kit Recommendation, with representative user requests covering Memory Read, Memory Write, and Memory Delete.}
    \label{fig:eval}
\end{figure*}

\begin{tcolorbox}[aclbox, width=\linewidth]
\textbf{ACL Trip Example (cont.).}
Consider the \textsc{Semantic XPath} query $Q_1$
:
\begin{center}
\path{//Day[ avg( POI[ node ~= "conference" ] ) ]}
\end{center}
$Q_1$ is parsed as a single-step query with an empty suffix, $Q_1=S\,\varnothing$, where the step 
 $S=ANP$ with $A=\texttt{//}$, $N=\textit{Day}$, $P={avg(S')}$, and $S'=$\path{POI[ node ~= "conference" ]}. Execution then proceeds as follows (\Autoref{fig:pipeline}):
\begin{enumerate}[leftmargin=*, itemsep=0pt, topsep=0pt, parsep=0pt, partopsep=0pt]
\item \textbf{Initialize:} Let $W_0=\{(r,1)\}$. Initialize execution as
$\mathsf{E}(Q,W_0).$

\item \textbf{Reduce the query:}
$
\mathsf{E}(Q,W_0)=\mathsf{E}(S,W_0)
$

\item \textbf{Apply axis navigator:} Let $W_1=\mathsf{E}(\texttt{//},W_0)$
\[
\begin{aligned}
\mathsf{E}(Q,W_0)
&=\mathsf{E}\!\big(NP,\,\mathsf{E}(\texttt{//},W_0)\big)\\
&=\mathsf{E}(NP,W_1).
\end{aligned}
\]
Here, axis navigator maps the root node to all its descendant nodes.

\item \textbf{Apply node selector:} Let $W_2=\mathsf{E}(\textit{Day},W_1)$
\[
\begin{aligned}
\mathsf{E}(NP,W_1)
&=\mathsf{E}\!\big(P,\,\mathsf{E}(\textit{Day},W_1)\big)\\
&=\mathsf{E}(P,W_2).
\end{aligned}
\]
Here, the node selector restricts the root node's descendants to the \textit{Day} node only.

\item \textbf{Apply semantic relevance:}
\[
\begin{aligned}
\mathsf{E}(P,W_2)
&=\mathsf{E}\!\big(\textit{avg}(S'),W_2\big).
\end{aligned}
\]

\item \textbf{Evaluate inner recursion:} 

$\mathsf{E}\!\big(\textit{avg}(S'),W_2\big)$ yields POI-level relevance scores for each day $d$ (i.e., how conference-related its child POIs are). For \emph{Day~2}, the POIs receive semantic relevance scores of $0.603$, $0.482$, and $0.608$ for \path{POI[ node ~= "conference" ]}, whose average gives a day-level score of $0.565$.

\item \textbf{Terminate:} Recursion ends at the base case $\mathsf{E}(\varnothing,\cdot)$, selecting \emph{Day~2} with three conference activities as the top-ranked match.
\end{enumerate}
\end{tcolorbox}



\section{Evaluation}

We evaluate \textsc{Semantic XPath} against two baselines (in-context memory and flat RAG) on two metrics: (1) \textbf{average pass rate}, the fraction of test requests for which the system retrieves the correct structured data and produces an output that satisfies the ground-truth constraints, and (2) \textbf{token usage}, the total LLM token cost per request.

\begin{itemize}
    \item[RQ1:] In a \emph{single-turn} setting with a static structured memory snapshot, how does \textsc{Semantic XPath} compare against existing methods on various ConvAI tasks (e.g., \Autoref{fig:eval})? 
     \item[RQ2:] In a \emph{multi-turn} setting that requires dynamic memory maintenance, how does each method perform on interactions that require retrieving revision history across versions, and how does the number of turns affect token usage?
\end{itemize}

\subsection{Setup}

\paragraph{Evaluation Domains.}
We consider three ConvAI scenarios involving structured long-term memory, with a simplified structure for each in \Autoref{fig:eval}.

\begin{itemize}
\item \textbf{Travel itinerary}: Seven-day travel plan, with each day consisting of restaurant and point-of-interest options annotated with descriptions, estimated costs, and preferences.
\item \textbf{To-do list}: Structured to-do list with task organization spanning 5 categories such as work, household, and personal categories. Each category contains related projects consisting of tasks annotated with a description, progress status, priority level, and deadline.
\item \textbf{Meal kit recommendation}: 7-day meal plans for a family, where each member is offered three options per meal time, each of which is annotated with ingredients and nutrition.
\end{itemize}

Each domain has a domain-specific schema and a curated interaction set consisting of 20 \emph{single-turn} conversations and 5 \emph{multi-turn} conversations. 

    

\paragraph{Configurations.}
We evaluate all methods using \texttt{GPT-5 mini} and \texttt{Gemini-3 flash} as LLM backbones. For {flat RAG}, we use the \texttt{Qwen3-Embedding-8B} embedding model. 

For \textsc{Semantic XPath}, we consider two semantic relevance scorers:
(1) the \emph{semantic similarity-based scorer}, which computes cosine similarity between the node content and the semantic relevance condition using \texttt{Qwen3-Embedding-8B} embeddings,
and (2) the \emph{entailment-based scorer}, which estimates the probability that the node content entails the relevance condition using \texttt{facebook/bart-large-mnli}.



\subsection{Evaluation Results}
\begin{figure}
    \centering
    \includegraphics[width=0.95\linewidth]{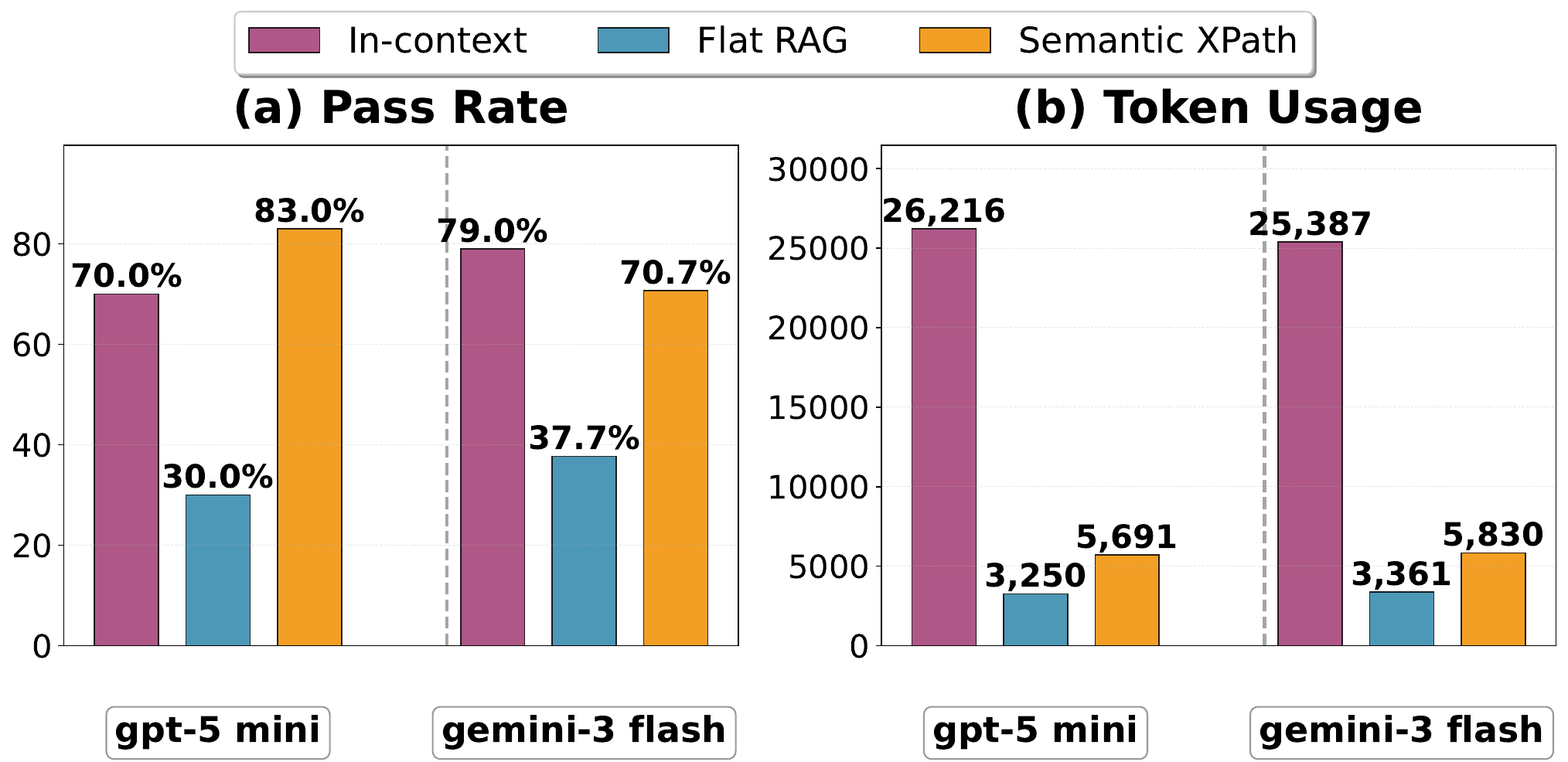}
    \caption{Single-turn evaluation across three methods. We report average pass rate (\textbf{left}) and token usage (\textbf{right}) averaged over three domains.}
    \label{fig:crud_results}
\end{figure}

\paragraph{RQ1: single-turn evaluation.}
We report single-turn conversation results averaged across three domains (\Autoref{fig:crud_results}, see \Autoref{app:results} for domain-specific results). As shown on the left of \Autoref{fig:crud_results}, \textsc{Semantic XPath} achieves a comparable pass rate to the in-context method, with both significantly outperforming the flat RAG baseline.

Flat RAG tends to return surface-level semantic matches that lack hierarchical reasoning. For example, for the request ``\textit{The weather forecast shows heavy rain on Day 7. Which activities are outdoors?}'', it retrieves outdoor POI across the entire trip rather than scoping results to Day 7.

The in-context method requires roughly 5$\times$ higher token usage than the other two methods (\Autoref{fig:crud_results}, right). Therefore, we can conclude that \textsc{Semantic XPath} achieves comparable effectiveness to the in-context baseline while requiring substantially fewer tokens.

\begin{figure}
    \centering
    \includegraphics[width=0.95\linewidth]{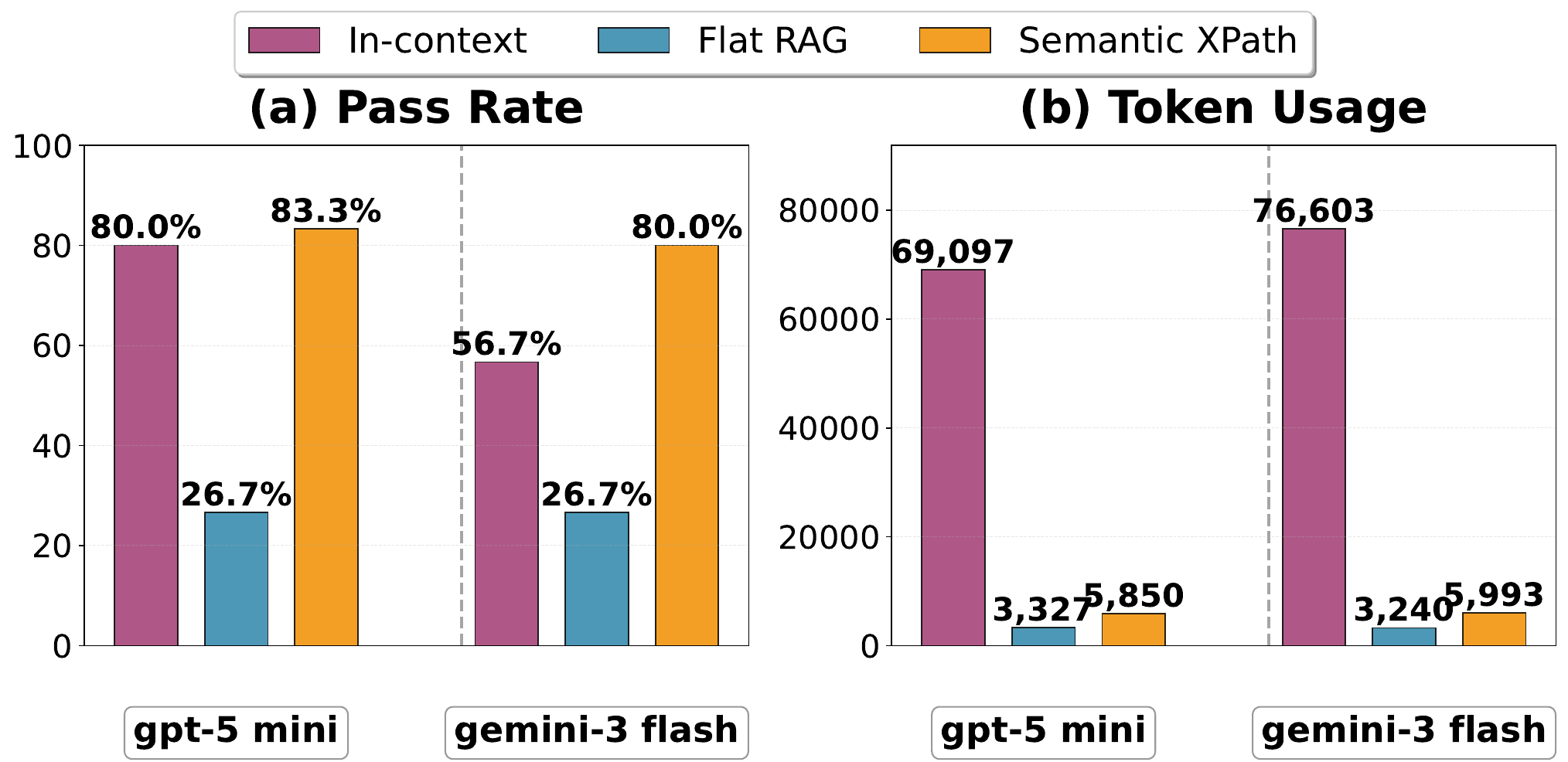}
    \caption{Multi-turn evaluation across three methods. We report average pass rate (\textbf{left}) and token usage (\textbf{right}) averaged over three domains.}
    \label{fig:version_results}
\end{figure}

\paragraph{RQ2: Multi-turn evaluation.} In \Autoref{fig:version_results}, the in-context baseline shows a drop in pass rate on requests that require accessing prior conversation history, as longer and more complex contexts increase the chance of misidentifying relevant information. In contrast, \textsc{Semantic XPath} retrieves only the relevant structured data, maintaining pass rates close to the single-turn setting.

\begin{figure}[t]
    \centering
    \includegraphics[width=\linewidth]{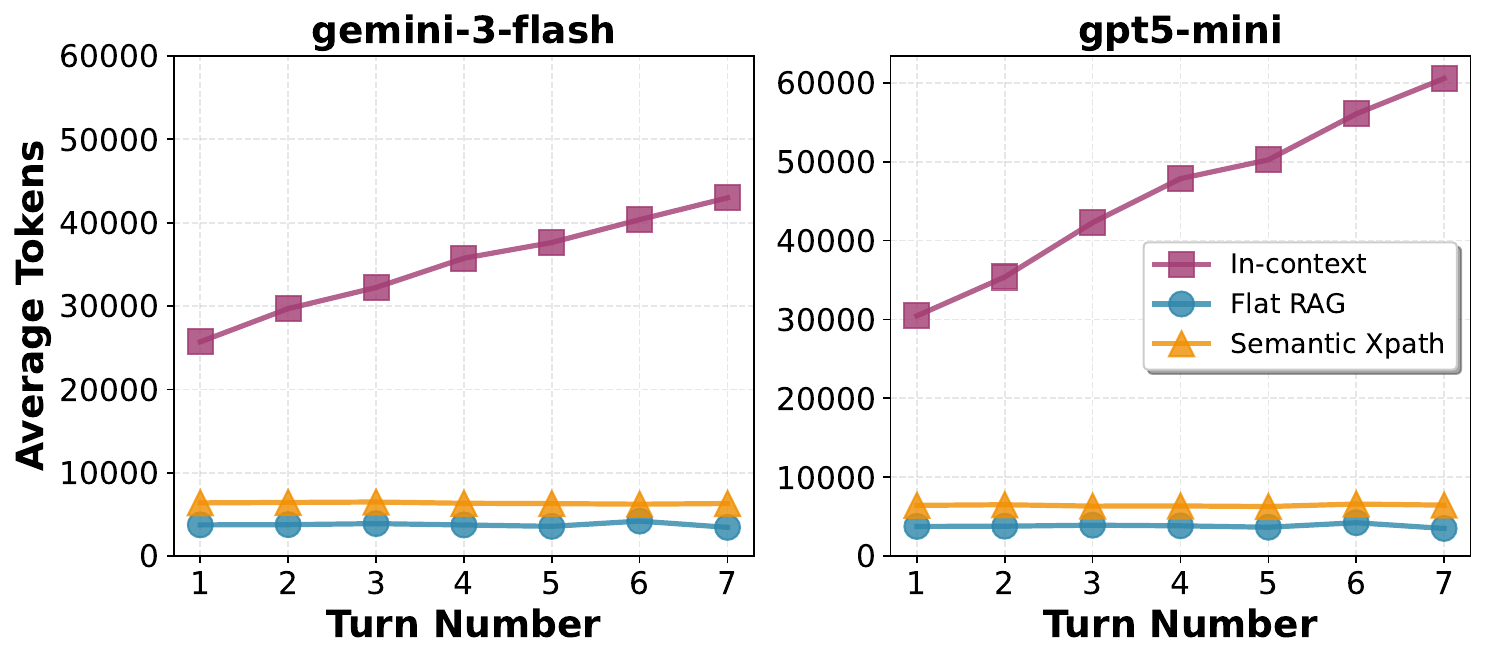}
\caption{Token usage vs.\ number of turns in a multi-turn ConvAI setting. Our proposed \textsc{Semantic XPath} maintains stable token consumption across turns, whereas the in-context baseline exhibits steadily increasing token usage as the conversation history grows.}
    \label{fig:multi-turn-token}
\end{figure}

In \Autoref{fig:multi-turn-token}, we evaluate token usage across turns. The in-context method shows a steady increase as it appends each user request and agent response to the input at every turn, whereas \textsc{Semantic XPath} maintains stable token usage by filtering out information irrelevant to the current request.

\section{Related Work}

\paragraph{External corpus vs. conversational memory.}
Prior work on memory for LLM agents often treats memory as an \emph{external corpus}, namely a static collection of documents or passages that exists independently of the conversation \citep{raptor, memwalker, cam, lattice, eqr1, eqr2, gpr}. This setting differs from \emph{conversational memory}, where memory represents the agent’s current working state for a task and is updated through interaction, making it inherently dynamic and mutable \citep{hmem, memtree}. Our focus is the conversational memory.

\paragraph{Flat vs. structured conversation memory.}
Within conversational memory, early approaches primarily store interaction history as \emph{flat} textual records with lightweight formatting or compressed summaries of past interactions \citep{memorybank, recursive_mem, memgpt, memo}. More recent work introduces \emph{structured} memory representations that organize information into hierarchies to support interaction across multiple abstraction levels \citep{raptor, memwalker, memtree, hmem, cam, lattice}. Our work focuses on structured conversational memory.

\paragraph{Abstractive vs. compositional hierarchies in structured memory.}
Structured conversational memory can be organized as an \emph{abstractive} hierarchy, where higher-level units are lossy summaries of lower-level content, or as a \emph{compositional} hierarchy
that structurally organizes task-relevant information 
(e.g., itinerary $\rightarrow$ day $\rightarrow$ activity). 
While most prior methods focus on abstractive hierarchies \citep{raptor, memwalker, memtree, hmem, cam, lattice}, compositional hierarchies are often more natural in task-oriented ConvAI because they reflect the hierarchical representation of interaction and tasks, with each level storing distinct information (e.g., an itinerary vs. individual activities).
We thus focus on \textbf{compositional hierarchies}.

\section{Conclusion}
This paper proposes \textsc{Semantic XPath}, a tree-structured memory module for ConvAI agents. Experiments show that \textsc{Semantic XPath} outperforms flat RAG baselines while using substantially fewer tokens than in-context memory. We also introduce \textsc{SemanticXPath Chat}, a live demo for \textsc{Semantic XPath} that visualizes structured memory and query execution details. Beyond providing an engaging long-term user experience for task-oriented settings such as itinerary planning, \textsc{Semantic XPath} serves as a practical framework for ConvAI developers to build more efficient structure-aware memory systems.

\section*{Ethical Considerations}

Our proposed method, \textsc{Semantic Xpath}, is a tree-structured memory module for ConvAI systems. As with any LLM-based approach, the system may inherit biases, factual inaccuracies, or reasoning errors from the underlying LLM. Furthermore, while the structured retrieval approach adopted in our work improves efficiency and relevance, it does not guarantee factual correctness. Users should therefore avoid over-reliance on generated outputs in high-stakes or safety-critical contexts.

\bibliography{references}
\newpage
\appendix

\section{Query Execution}
\label{app:eval}

\subsection{Step Evaluation}
\[
\mathsf{E}(S,W)
=
\begin{cases}
\mathsf{E}\!\big(NRP,\,\mathsf{E}(A,W)\big),
& \text{if }\substack{S=\\ANRP},\\[6pt]

\mathsf{E}\!\big(NP,\,\mathsf{E}(A,W)\big),
& \text{if }\substack{S=\\ANP},\\[6pt]

\mathsf{E}\!\big(NR,\,\mathsf{E}(A,W)\big),
& \text{if }\substack{S=\\ANR},\\[6pt]

\mathsf{E}\!\big(N,\,\mathsf{E}(A,W)\big),
& \text{if }\substack{S=\\AN}.
\end{cases}
\]

\subsection{Axis Selector Evaluation}

Define the axis expansion operators:
\[
\begin{aligned}
\mathsf{E}(\texttt{/},W)
&=\Big\{(u,w)\ \Big|\ \substack{\exists (v,w)\in W,\\ u\in \mathsf{C}(v)}\Big\},\\
\mathsf{E}(\texttt{//},W)
&=\Big\{(u,w)\ \Big|\ \substack{\exists (v,w)\in W,\\ u\in \mathsf{D}(v)}\Big\}.
\end{aligned}
\]
where $\mathsf{C}(v)$ and $\mathsf{D}(v)$ denote the child and descendant sets of node $v$, respectively.

The evaluation function for the axis selector is then defined as:
\[
\mathsf{E}(A,W)
=
\begin{cases}
\mathsf{E}(\texttt{/},W), & \text{if } A=\texttt{/},\\[4pt]
\mathsf{E}(\texttt{//},W), & \text{if } A=\texttt{//}.
\end{cases}
\]

\subsection{Node Selector Evaluation}

Define the node selector operators:
\[
\begin{aligned}
\mathsf{E}(\texttt{NodeType},W)
&=\Big\{(u,w)\ \Big|\ \substack{(u,w)\in W,\\ \kappa(u)=\texttt{NodeType}}\Big\},\\
\mathsf{E}(\texttt{*},W)
&=W.
\end{aligned}
\]
where $\kappa(u)$ returns the node type of $u$.

The evaluation function for the node selector is then defined as:
\[
\mathsf{E}(N,W)
=
\begin{cases}
\mathsf{E}(\texttt{NodeType},W), & \text{if }\substack{N=\\\texttt{NodeType}},\\[6pt]
\mathsf{E}(\texttt{*},W), & \text{if } N=\texttt{*}.
\end{cases}
\]

\subsection{Positional Selector Evaluation}

Define the positional selector operators:
\[
\begin{aligned}
\mathsf{E}([i],W)
&=[i](W),\\
\mathsf{E}([-i],W)
&=[-i](W),\\
\mathsf{E}([i{:}j],W)
&=[i{:}j](W),
\end{aligned}
\]
where $[i](\cdot)$, $[-i](\cdot)$, and $[i{:}j](\cdot)$ apply positional selection to $W$ in the topological order.

The evaluation function for the positional selector is then defined as:
\[
\mathsf{E}(R,W)
=
\begin{cases}
\mathsf{E}([i],W), & \text{if }R=[i],\\[6pt]
\mathsf{E}([-i],W), & \text{if } R=[-i],\\[6pt]
\mathsf{E}([i{:}j],W), & \text{if } R=[i{:}j].
\end{cases}
\]

\subsection{Semantic Relevance Operator Evaluation}

We define a semantic relevance scoring function
$
\mathsf{Rel}: V \times \Phi \rightarrow [0,1],
$
which assigns a graded relevance score to a node $u$ under a semantic relevance condition
$P\in\Phi$.

\paragraph{Local semantic relevance.}
\label{sec:local_pred}
Define the relevance scoring function for local semantic relevance:
\[
\mathsf{Rel}(u,\texttt{Local})
=
\mathsf{Atom}_{\mathrm{loc}}(u,\texttt{Local}),
\]
where $\mathsf{Atom}_{\mathrm{loc}}(u,\texttt{Local})\in[0,1]$ is computed from either a specific attribute
or the full node representation via a semantic scoring function.

\paragraph{Aggregation semantic relevance.}
\label{sec:agg_pred}
If $P := \mathsf{Agg}(S')$, where the inner step $S'$ has the surface form
$S' := N\;R\;P'$, let $\mathcal{E}_{S'}^{(u)} \subseteq V$ denote the evidence node set obtained by
structurally executing $S'$ from $u$. The aggregation relevance scoring function is defined recursively as
\[
\mathsf{Rel}(u,\mathsf{Agg}(S'))
=
\mathsf{Agg}
\Big(
\{\, \mathsf{Rel}(x,P') \mid x \in \mathcal{E}_{S'}^{(u)} \,\}
\Big),
\]
where $\mathsf{Agg}$ is an aggregation operator such as \texttt{min}, \texttt{max}, \texttt{avg}, or \texttt{gmean}.

The evaluation function for the semantic relevance operator is then defined as:
\[
\mathsf{E}(P,W)
=
\{(u,\; w\cdot \mathsf{Rel}(u,P)) \mid (u,w)\in W\}.
\]

\paragraph{Compositional semantic relevance.}
For the unary operator $1-P$, we define
\[
\mathsf{Rel}(u,\,1-P)=1-\mathsf{Rel}(u,P),
\]
and the corresponding semantic relevance operator
\[
\mathsf{E}(1-P,W)
=
\Big\{(u,\; w\cdot \mathsf{Rel}(u,1-P))\ \Big|\ \substack{(u,w)\in W}\Big\}.
\]

For binary compositional conditions, let $\odot \in \{\textit{avg},\textit{prod},\textit{min},\textit{max}\}$ be a binary operator on $[0,1]$,
and for any node $u$ and relevance conditions $P_1,P_2$, define
\[
\mathsf{Rel}(u,\,P_1 \odot P_2)=\odot\big(\mathsf{Rel}(u,P_1),\,\mathsf{Rel}(u,P_2)\big).
\]
The corresponding compositional semantic relevance operator is
\[
\mathsf{E}(P_1\odot P_2,W)
=
\Big\{(u,\; w\cdot \mathsf{Rel}(u,P_1\odot P_2))\ \Big|\ \substack{(u,w)\in W}\Big\}.
\]

\begin{table*}[htbp]
\centering
\small
\setlength{\tabcolsep}{4pt}
\begin{tabular}{@{\hspace{4pt}}lllcc@{\hspace{6pt}}cc@{\hspace{6pt}}cc@{\hspace{6pt}}cc@{\hspace{4pt}}}
\toprule
Model & Method & Scorer
& \multicolumn{2}{c}{\colorbox{seclight}{\strut\hspace{3pt}Itinerary\hspace{3pt}}}
& \multicolumn{2}{c}{\colorbox{seclight}{\strut\hspace{3pt}To Do List\hspace{3pt}}}
& \multicolumn{2}{c}{\colorbox{seclight}{\strut\hspace{3pt}Meal Kit\hspace{3pt}}}
& \multicolumn{2}{c}{\colorbox{seclight}{\strut\hspace{3pt}Avg.\hspace{3pt}}} \\
\cmidrule(lr){4-5}\cmidrule(lr){6-7}\cmidrule(lr){8-9}\cmidrule(lr){10-11}
 &  &  & Pass & Tokens & Pass & Tokens & Pass & Tokens & Pass & Tokens \\
\midrule

\multirow{4}{*}{\centering GPT-5 mini}
& In-context     & --     & 55.0 & 12,136 & 80.0 & 5,444 & 75.0 & 61,069 & 70.0 & 26,216 \\
& Flat RAG       & --     & 25.0 & 3,137  & 40.0 & 2,722 & 25.0 & 3,892  & 30.0 & 3,250 \\
\rowcolor{sxblue}
\cellcolor{white}\multirow{-2}{*}{\centering GPT-5 mini}
& Semantic XPath & \textit{Sem}    & 75.0 & 5,548  & 64.0 & 5,937 & 60.0 & 7,013  & 66.3 & 6,166 \\
\rowcolor{sxblue}
\cellcolor{white}
& Semantic XPath & \textit{Entail} & 85.0 & 5,161  & 84.0 & 5,184 & 80.0 & 6,728  & 83.0 & 5,691 \\

\midrule

\multirow{4}{*}{\centering Gemini 3 Flash}
& In-context     & --     & 95.0 & 12,846 & 72.0 & 10,465 & 70.0 & 52,849 & 79.0 & 25,387 \\
& Flat RAG       & --     & 35.0 & 3,349  & 48.0 & 2,934  & 30.0 & 3,800  & 37.7 & 3,361 \\
\rowcolor{sxblue}
\cellcolor{white}\multirow{-2}{*}{\centering Gemini 3 Flash}
& Semantic XPath & \textit{Sem}    & 65.0 & 5,454  & 68.0 & 5,710  & 60.0 & 6,966  & 64.3 & 6,043 \\
\rowcolor{sxblue}
\cellcolor{white}
& Semantic XPath & \textit{Entail} & 75.0 & 5,240  & 72.0 & 5,605  & 65.0 & 6,644  & 70.7 & 5,830 \\

\bottomrule
\end{tabular}
\caption{Single-turn evaluation across methods. \textit{Sem} denotes the Semantic Similarity scorer and \textit{Entail} denotes the Entailment scorer for \textsc{Semantic XPath}. We report average pass rate and token usage with GPT-5 mini and Gemini 3 Flash as backbones across three domains: Travel Itinerary, To Do List, and Meal Kit Recommendation.}
\label{tab:crud_results_agg}
\end{table*}

\begin{table*}[h]
\centering
\small
\setlength{\tabcolsep}{4pt}
\begin{tabular}{@{\hspace{4pt}}lllcc@{\hspace{6pt}}cc@{\hspace{6pt}}cc@{\hspace{6pt}}cc@{\hspace{4pt}}}
\toprule
Model & Method & Scorer
& \multicolumn{2}{c}{\colorbox{seclight}{\strut\hspace{3pt}Itinerary\hspace{3pt}}}
& \multicolumn{2}{c}{\colorbox{seclight}{\strut\hspace{3pt}To Do List\hspace{3pt}}}
& \multicolumn{2}{c}{\colorbox{seclight}{\strut\hspace{3pt}Meal Kit\hspace{3pt}}}
& \multicolumn{2}{c}{\colorbox{seclight}{\strut\hspace{3pt}Avg.\hspace{3pt}}} \\
\cmidrule(lr){4-5}\cmidrule(lr){6-7}\cmidrule(lr){8-9}\cmidrule(lr){10-11}
 &  &  & Pass & Tokens & Pass & Tokens & Pass & Tokens & Pass & Tokens \\
\midrule

\multirow{4}{*}{\centering GPT-5 mini}
& In-context     & --     & 90.0 & 28,342 & 70.0 & 20,398 & 80.0 & 158,552 & 80.0 & 69,097 \\
& Flat RAG       & --     & 20.0 & 3,250  & 50.0 & 3,043  & 10.0 & 3,686   & 26.7 & 3,327 \\
\rowcolor{sxblue}
\cellcolor{white}\multirow{-2}{*}{\centering GPT-5 mini}
& Semantic XPath & \textit{Sem}    & 50.0 & 5,664  & 100.0 & 6,132 & 60.0 & 7,087   & 70.0 & 6,294 \\
\rowcolor{sxblue}
\cellcolor{white}
& Semantic XPath & \textit{Entail} & 80.0 & 5,820  & 100.0 & 5,649 & 70.0 & 6,079   & 83.3 & 5,850 \\

\midrule

\multirow{4}{*}{\centering Gemini 3 Flash}
& In-context     & --     & 60.0 & 31,639 & 70.0 & 23,602 & 40.0 & 174,567 & 56.7 & 76,603 \\
& Flat RAG       & --     & 20.0 & 3,164  & 50.0 & 2,959  & 10.0 & 3,596   & 26.7 & 3,240 \\
\rowcolor{sxblue}
\cellcolor{white}\multirow{-2}{*}{\centering Gemini 3 Flash}
& Semantic XPath & \textit{Sem}    & 60.0 & 5,451  & 100.0 & 5,872 & 70.0 & 6,954   & 76.7 & 6,092 \\
\rowcolor{sxblue}
\cellcolor{white}
& Semantic XPath & \textit{Entail} & 60.0 & 5,671  & 100.0 & 5,526 & 80.0 & 6,782   & 80.0 & 5,993 \\

\bottomrule
\end{tabular}
\caption{Multi-turn evaluation across methods. \textit{Sem} denotes the Semantic Similarity scorer and \textit{Entail} denotes the Entailment scorer for \textsc{Semantic XPath}. We report average pass rate and token usage with GPT-5 mini and Gemini 3 Flash as backbones across three domains: Travel Itinerary, To Do List, and Meal Kit Recommendation.}
\label{tab:version_results_agg}
\end{table*}

The evaluation function for the semantic relevance operator is then defined as:
\[
\mathsf{E}(P,W)
=
\begin{cases}
\mathsf{E}(\texttt{Local},W), & \text{if }\substack{P=\\\texttt{Local}},\\[6pt]
\mathsf{E}(\mathsf{Agg}(S'),W), & \text{if }\substack{P=\\\mathsf{Agg}(S')},\\[6pt]
\mathsf{E}(1-P',W), & \text{if }\substack{P=\\1-P'},\\[6pt]
\mathsf{E}(P_1\odot P_2,W), & \text{if }\substack{P=\\P_1\odot P_2},
\end{cases}
\]

\section{Additional Results}\label{app:results}

Across both single-turn and multi-turn results (\Autoref{tab:crud_results_agg} and \Autoref{tab:version_results_agg}), in-context memory achieves strong pass rates but incurs very high token usage, especially on meal kit recommendation queries. \textsc{Semantic XPath} substantially reduces token cost while maintaining competitive accuracy, and the entailment scorer consistently outperforms the semantic similarity scorer in pass rate for both GPT-5 mini and Gemini 3 Flash. Flat RAG remains the lowest-cost baseline but with noticeably lower pass rates.

\section{Case Study}

Beyond the single-turn requests shown in \Autoref{fig:pipeline}, a real ConvAI system must support long-term, task-oriented interactions. Users may revise a plan over many turns and later return to earlier versions or even different plans. The key challenge is whether the system can reliably retrieve the right information across revisions and plans despite ongoing edits and growing history.

Below, we demonstrate a case with example plan revision requests from users using the same ACL trip example:

\begin{tcolorbox}[aclbox, width=\linewidth]
\textbf{ACL Trip Example (cont.)}
We illustrate two requests that interact with revision history.

\smallskip
\noindent\textbf{Request 1:} \emph{``Cancel the poster session visit cause I need to take a client meeting.''}\\
\noindent\textbf{Query:}
\begin{center}
\path{//POI[ node ~= "poster" ]}
\end{center}

\smallskip
\noindent\textbf{10 follow-up requests omitted:} $\cdots$

\smallskip
\noindent\textbf{Request 2:} \emph{``Wait, what was the poster session time again? I might be able to make it.''}\\
\noindent\textbf{Query:}
\begin{center}
\path{//Version[node~= "delete poster session"]//POI[ node ~= "poster" ]}
\end{center}
\noindent\textbf{Result:} \textsc{Semantic XPath} retrieves the version created by the deletion edit, recovers the removed poster-session entry from that revision context, and returns its scheduled time. In contrast, in-context memory fails under the longer context.
\end{tcolorbox}

\end{document}